\documentclass{article}
\usepackage{arxiv}
\usepackage[utf8]{inputenc}
\usepackage[T1]{fontenc}
\usepackage{amsmath,amssymb,amsfonts}
\usepackage{graphicx}
\usepackage{booktabs}
\usepackage{hyperref}
\usepackage{cleveref}
\usepackage{xcolor}
\usepackage{algorithm}
\usepackage{algorithmic}
\usepackage{natbib}
\usepackage{doi}
\usepackage{tikz}
\usepackage{subcaption}
\usepackage{multirow}
\usetikzlibrary{positioning, arrows.meta, shapes.geometric, calc}
\usepackage{pgfplots}
\pgfplotsset{compat=1.18}

\definecolor{lowcolor}{RGB}{215, 48, 39}      
\definecolor{medcolor}{RGB}{253, 174, 97}     
\definecolor{highcolor}{RGB}{26, 152, 80}     

\newcommand{\sgi}{\text{SGI}}
\newcommand{\R}{\mathbb{R}}
\newcommand{\Sphere}{\mathbb{S}}

\newcommand{\embed}{\phi}

\title{Semantic Grounding Index: Geometric Bounds on Context Engagement in RAG Systems}

\author{
  Javier Marin\\
  CERT\\
  \texttt{javier@jmarin.info}
}

\begin{document}

\maketitle

\begin{abstract}
When retrieval-augmented generation (RAG) systems hallucinate, what geometric trace does this leave in embedding space? We introduce the Semantic Grounding Index (SGI), defined as the ratio of angular distances from the response to the question versus the context on the unit hypersphere $\Sphere^{d-1}$. Our central finding is \emph{semantic laziness}: hallucinated responses remain angularly proximate to questions rather than departing toward retrieved contexts. On HaluEval ($n=5{,}000$), we observe large effect sizes (Cohen's $d$ ranging from $0.92$ to $1.28$) across five embedding models with mean cross-model correlation $r = 0.85$. Crucially, we derive from the spherical triangle inequality that SGI's discriminative power should increase with question-context angular separation $\theta(q,c)$---a theoretical prediction confirmed empirically: effect size rises monotonically from $d = 0.61$ (low $\theta(q,c)$) to $d = 1.27$ (high $\theta(q,c)$), with AUC improving from $0.72$ to $0.83$. Subgroup analysis reveals that SGI excels on long responses ($d = 2.05$) and short questions ($d = 1.22$), while remaining robust across context lengths. Calibration analysis yields ECE $= 0.10$, indicating SGI scores can serve as probability estimates, not merely rankings. A critical negative result on TruthfulQA (AUC $= 0.478$) establishes that angular geometry measures topical engagement rather than factual accuracy. SGI provides computationally efficient, theoretically grounded infrastructure for identifying responses that warrant verification in production RAG deployments.
\end{abstract}

\section{Introduction}

LLMs generate text through autoregressive next-token prediction, optimizing for distributional statistics of training corpora \citep{brown2020language,touvron2023llama}. This objective produces fluent continuations without maintaining explicit correspondence to external reality, a characteristic that manifests as \emph{hallucination} \citep{ji2023survey,huang2023survey}. Retrieval-Augmented Generation (RAG) architectures condition generation on retrieved documents \citep{lewis2020retrieval,guu2020realm}, yet hallucination persists: models fabricate claims absent from context or fail to substantively engage with retrieved information \citep{shuster2021retrieval,bao2025faithbench,niu2024ragtruth}.

We investigate a geometric question: when a RAG system fails to ground its response in the provided context, what signature does this leave in embedding space? Modern sentence transformers are trained via contrastive objectives that explicitly optimize angular relationships on the unit hypersphere \citep{reimers2019sentence,wang2020understanding}. This makes $\Sphere^{d-1}$ the natural geometric setting for analyzing response behavior.

Our central contribution is the identification, theoretical characterization, and empirical validation of a geometric pattern we term \emph{semantic laziness}. When models hallucinate in RAG systems, their responses remain angularly proximate to the question rather than departing toward the context's semantic territory. We formalize this through the Semantic Grounding Index (SGI)---the ratio of angular distances $\theta(r,q)/\theta(r,c)$---and demonstrate that it provides a robust, theoretically grounded signal for detecting context disengagement.

\paragraph{Contributions.}
\begin{enumerate}
    \item We introduce SGI as an intrinsic quantity on $\Sphere^{d-1}$ and derive geometric bounds from the spherical triangle inequality that \emph{predict} when discrimination will be most effective.
    
    \item We \emph{confirm this theoretical prediction empirically}: effect size increases monotonically with question-context angular separation ($d = 0.61 \rightarrow 0.90 \rightarrow 1.27$ across $\theta(q,c)$ terciles).
    
    \item We establish cross-model robustness at scale ($n = 5{,}000$): five architecturally distinct embedding models correlate at $r = 0.85$ with ranking agreement $\rho = 0.87$.
    
    \item We characterize operational boundaries: SGI excels on long responses and short questions, maintains calibration (ECE $= 0.10$), and fails predictably on TruthfulQA where angular geometry cannot discriminate factual accuracy.
\end{enumerate}

\section{Theoretical Foundations}

\subsection{The Embedding Hypersphere}

Contrastive learning objectives for sentence embeddings decompose into alignment and uniformity terms on the unit hypersphere \citep{wang2020understanding}. The InfoNCE loss encourages matched pairs to cluster while spreading unrelated points apart, inducing structure on $\Sphere^{d-1}$ where L2-normalized embeddings reside.

Let $\embed: \mathcal{S} \rightarrow \R^d$ denote a sentence embedding model and $\hat{\embed}(s) = \embed(s)/\|\embed(s)\|$ the L2-normalized representation. The normalized embeddings lie on:
\begin{equation}
\Sphere^{d-1} = \{x \in \R^d : \|x\| = 1\}
\end{equation}

This is a compact Riemannian manifold with constant positive curvature \citep{mardia2000directional}. The intrinsic distance is the geodesic (great-circle arc length):
\begin{equation}
\theta(a, b) = \arccos(\hat{\embed}(a)^\top \hat{\embed}(b))
\label{eq:geodesic}
\end{equation}

This angular distance $\theta \in [0, \pi]$ satisfies all metric axioms on $\Sphere^{d-1}$, including the triangle inequality \citep{bridson2013metric}. We note that while cosine similarity is ubiquitous in applications, it does not satisfy the triangle inequality; angular distance is the proper metric for geometric analysis \citep{you2025semantics}.

\subsection{The Semantic Grounding Index}

For a RAG instance $(q, c, r)$ with question $q$, retrieved context $c$, and generated response $r$, we define the Semantic Grounding Index as:
\begin{equation}
\sgi(r; q, c) = \frac{\theta(r, q)}{\theta(r, c)} = \frac{\arccos(\hat{\embed}(r)^\top \hat{\embed}(q))}{\arccos(\hat{\embed}(r)^\top \hat{\embed}(c))}
\label{eq:sgi}
\end{equation}

Equation \ref{eq:sgi} measures the ratio of angular departures, how far the response has traveled from the question relative to its distance from the context. When $\sgi > 1$, the response is angularly farther from the question than from the context---it has ``departed'' toward the context's semantic territory. When $\sgi < 1$, the response remains closer to the question than to the context.

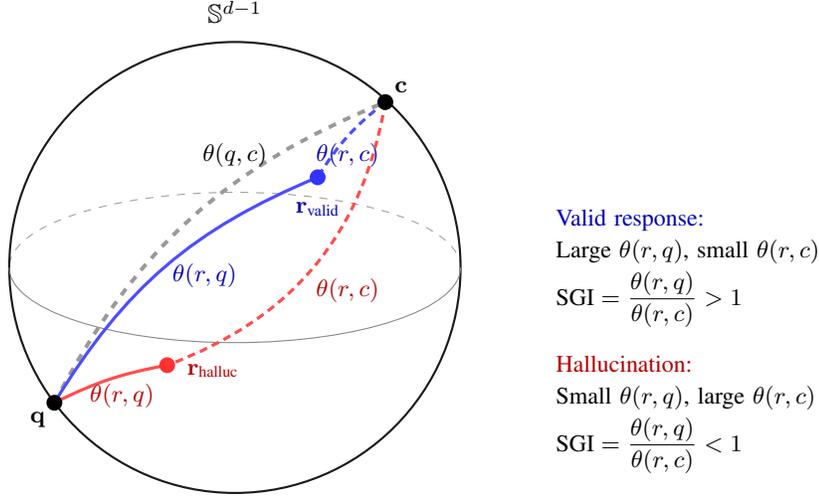
\begin{figure}[t]
\centering
\begin{tikzpicture}[scale=2]
    \def\R{1.5}
    
    \draw[gray!75, dashed] (\R,0) arc (0:180:\R cm and 0.5cm);
    
    \draw[thick, black!90] (0,0) circle (\R);
    
    \draw[gray!95] (-\R,0) arc (180:360:\R cm and 0.5cm);
    
    \coordinate (q) at (-1.2, -0.9);
    \coordinate (c) at (1.0, 1.1);
    \coordinate (rv) at (0.55, 0.6);
    \coordinate (rh) at (-0.45, -0.65);
    
    \draw[black!40, line width=1.5pt, dashed] (q) to[bend left=20] (c);
    \node[font=\footnotesize, black!80!black] at (0, 0.75) {$\theta(q,c)$};
    
    \draw[blue!70, line width=1.2pt] (q) to[bend left=18] (rv);
    \node[font=\footnotesize, blue!70!black] at (-0.2, -0.05) {$\theta(r,q)$};
    
    \draw[blue!70, line width=1.2pt, densely dashed] (c) to[bend right=10] (rv);
    \node[font=\footnotesize, blue!70!black] at (0.75, 0.75) {$\theta(r,c)$};
    
    \draw[red!70, line width=1.2pt] (q) to[bend left=8] (rh);
    \node[font=\footnotesize, red!70!black] at (-0.75, -0.85) {$\theta(r,q)$};
    
    \draw[red!70, line width=1.2pt, densely dashed] (c) to[bend left=30] (rh);
    \node[font=\footnotesize, red!70!black] at (0.75, -0.15) {$\theta(r,c)$};
    
    \fill[black] (q) circle (1.5pt);
    \node[below left, font=\small, black] at (q) {$\mathbf{q}$};
    
    \fill[black] (c) circle (1.5pt);
    \node[above right, font=\small, black] at (c) {$\mathbf{c}$};
    
    \fill[blue!80] (rv) circle (1.5pt);
    \node[above, font=\small, blue!70!black, yshift=-18pt] at (rv) {$\mathbf{r}_{\text{valid}}$};
    
    \fill[red!80] (rh) circle (1.5pt);
    \node[right, font=\small, red!70!black, xshift=4pt, yshift=-2pt] at (rh) {$\mathbf{r}_{\text{halluc}}$};
    
    \node[font=\normalsize] at (0, 1.7) {$\mathbb{S}^{d-1}$};
    
\end{tikzpicture}
\hspace{0.8cm}
\vspace{0.5cm}
\begin{tikzpicture}[scale=0.7]
    \node[font=\small, align=left, 
          fill=gray!0, text width=4.5cm, inner sep=8pt] at (2,2) {
        \textcolor{blue!70!black}{Valid response:}\\[2pt]
        Large $\theta(r,q)$, small $\theta(r,c)$\\[2pt]
        $\displaystyle\text{SGI} = \frac{\theta(r,q)}{\theta(r,c)} > 1$\\[10pt]
        \textcolor{red!70!black}{Hallucination:}\\[2pt]
        Small $\theta(r,q)$, large $\theta(r,c)$\\[2pt]
        $\displaystyle\text{SGI} = \frac{\theta(r,q)}{\theta(r,c)} < 1$
    };
\end{tikzpicture}

\caption{Angular geometry of SGI on the unit hypersphere. Question $\mathbf{q}$ and context $\mathbf{c}$ define anchor points; their angular separation $\theta(q,c)$ determines the geometric ``room'' for response differentiation. A valid response (blue) departs from $\mathbf{q}$ toward $\mathbf{c}$, yielding SGI $> 1$. A hallucination (red) remains angularly proximate to the question---the semantic laziness signature---yielding SGI $< 1$.}
\label{fig:sphere}
\end{figure}

\subsection{Geometric Bounds and Theoretical Predictions}

The spherical triangle inequality constrains admissible SGI values. For any $q, c, r \in \Sphere^{d-1}$:
\begin{equation}
|\theta(q, c) - \theta(r, c)| \leq \theta(r, q) \leq \theta(q, c) + \theta(r, c)
\end{equation}

Dividing by $\theta(r, c)$ yields bounds on SGI:
\begin{equation}
\left|\frac{\theta(q, c)}{\theta(r, c)} - 1\right| \leq \sgi \leq \frac{\theta(q, c)}{\theta(r, c)} + 1
\label{eq:bounds}
\end{equation}

These bounds form the basis of our theoretical contribution:

\begin{quote}
\textit{SGI's discriminative power should increase with $\theta(q,c)$. When question and context are semantically similar (small ($\theta(q,c)$)), the triangle inequality constrains SGI values near 1 regardless of response quality. When $\theta(q,c)$ is large, the constraint relaxes, permitting greater separation between grounded and ungrounded responses.}
\end{quote}

This is the mathematical consequence of the triangle inequality. If SGI captures something real about semantic grounding, we should observe effect sizes that increase monotonically with $\theta(q,c)$. We test this prediction explicitly in Section~\ref{sec:stratified}.

\subsection{The Semantic Laziness Hypothesis}

We hypothesize that hallucinated responses in RAG systems exhibit \emph{semantic laziness}: rather than introducing vocabulary and concepts from the retrieved context, they produce completions that remain in the question's semantic neighborhood.

Let $\mathcal{R}_{\text{valid}}$ and $\mathcal{R}_{\text{halluc}}$ denote distributions over valid and hallucinated responses. The semantic laziness hypothesis predicts:
\begin{equation}
\mathbb{E}_{r \sim \mathcal{R}_{\text{valid}}}[\sgi(r; q, c)] > \mathbb{E}_{r \sim \mathcal{R}_{\text{halluc}}}[\sgi(r; q, c)]
\label{eq:hypothesis}
\end{equation}

This hypothesis connects to how autoregressive models handle uncertainty. When a model lacks confidence in how to use retrieved context, it may default to ``safe'' completions that echo the question's framing rather than venturing into the context's semantic territory.

\section{Related Work}

\subsection{Geometric Methods for Hallucination Detection}
\citet{li2025semantic} compute semantic volume from batches of responses to quantify uncertainty. \citet{catak2024uncertainty} apply convex hull analysis to embedding spaces. \citet{gao2025attention} found that hallucinated responses exhibit smaller deviations from prompts in hidden state space---an observation consistent with our semantic laziness characterization. Our work differs by focusing on the triangular geometry of question-context-response relationships, deriving theoretical bounds, and establishing cross-model robustness.

\subsection{Semantic Entropy and Consistency Methods}
\citet{farquhar2024semantic} introduced semantic entropy for hallucination detection via multiple sampling. \citet{kuhn2023semantic} developed linguistic invariances for uncertainty estimation. These methods require multiple generation passes; SGI operates on single responses.

\subsection{NLI-Based Detection}
SummaC \citep{laban2022summac}, HALT-RAG \citep{halt2025}, and LettuceDetect \citep{kovacs2025lettucedetect} frame detection as entailment classification. These methods detect logical contradiction; SGI detects semantic disengagement. The signals are complementary.

\subsection{Spherical Geometry in Representation Learning}
The unit hypersphere is well-studied in directional statistics \citep{mardia2000directional,fisher1953dispersion}. \citet{wang2020understanding} analyzed contrastive learning through alignment and uniformity on $\Sphere^{d-1}$. \citet{meng2019spherical} developed spherical text embeddings. \citet{you2025semantics} provided comprehensive analysis of when cosine similarity succeeds and fails, noting that angular distance---unlike cosine similarity---satisfies the triangle inequality.

\section{Experimental Design}

\subsection{Implementation Details}

\paragraph{Text Preprocessing.}
We use spaCy \citep{spacy} with the \texttt{en\_core\_web\_sm} pipeline for sentence boundary detection and basic tokenization when segmenting long contexts. This lightweight model (12MB) provides sufficient accuracy for our preprocessing needs without introducing computational overhead. Alternative pipelines include \texttt{en\_core\_web\_md} (40MB) and \texttt{en\_core\_web\_lg} (560MB), which incorporate word vectors but offer no advantage for boundary detection. For purely rule-based segmentation, spaCy's \texttt{sentencizer} component or NLTK's \texttt{punkt} tokenizer are viable alternatives; we observed no significant difference in downstream SGI scores across these choices, suggesting robustness to preprocessing variations.

\paragraph{Embedding Computation.}
We compute sentence embeddings using the \texttt{sentence-transformers} library (v2.2.2) \citep{reimers2019sentence}. For each RAG instance $(q, c, r)$, we encode the question, context, and response as separate strings without additional prompting or instruction prefixes. Embeddings are L2-normalized to unit length before angular distance computation.

\paragraph{SGI Computation.}
Algorithm~\ref{alg:sgi} specifies the complete procedure. Angular distances are computed via $\theta(a,b) = \arccos(\text{clip}(\mathbf{a}^\top \mathbf{b}, -1, 1))$, where clipping prevents numerical errors from domain violations. We add $\epsilon = 10^{-8}$ to the denominator to avoid division by zero when $\theta(r,c) \approx 0$.

\begin{center}
\begin{minipage}{0.60\textwidth}
\begin{algorithm}[H]
\caption{Semantic Grounding Index Computation}
\label{alg:sgi}
\begin{algorithmic}[1]
\REQUIRE Question $q$, Context $c$, Response $r$, Embedding model $\phi$
\ENSURE SGI score
\STATE $\mathbf{q} \leftarrow \phi(q) / \|\phi(q)\|$ \COMMENT{L2 normalize}
\STATE $\mathbf{c} \leftarrow \phi(c) / \|\phi(c)\|$
\STATE $\mathbf{r} \leftarrow \phi(r) / \|\phi(r)\|$
\STATE $\theta_{rq} \leftarrow \arccos(\text{clip}(\mathbf{r}^\top \mathbf{q}, -1, 1))$
\STATE $\theta_{rc} \leftarrow \arccos(\text{clip}(\mathbf{r}^\top \mathbf{c}, -1, 1))$
\STATE \textbf{return} $\theta_{rq} / (\theta_{rc} + \epsilon)$
\end{algorithmic}
\end{algorithm}
\end{minipage}
\end{center}

\paragraph{Sampling and Splits.}
From HaluEval QA ($10{,}000$ samples), we randomly sample $n = 5{,}000$ instances stratified by hallucination label. For TruthfulQA, we use all 817 questions, constructing paired samples by treating each question's correct and incorrect answers as separate instances ($n = 800$ after filtering incomplete entries). No train/test split is applied; we report descriptive statistics on the full samples.

\paragraph{Statistical Analysis.}
Effect sizes use Cohen's $d$ with pooled standard deviation. Group comparisons use Welch's $t$-test (unequal variances). Cross-model correlations use Pearson $r$ for linear agreement and Spearman $\rho$ for rank agreement. Calibration analysis uses expected calibration error (ECE) with 10 equal-frequency bins. 

\subsection{Datasets}

\paragraph{HaluEval QA} \citep{li2023halueval} provides question-knowledge-answer triples with hallucination labels. The knowledge field serves as retrieved context. We use $n = 5{,}000$ samples for comprehensive analysis, enabling stratified evaluation with sufficient statistical power.

\paragraph{TruthfulQA} \citep{lin2022truthfulqa} contains 817 questions targeting common misconceptions, with truthful and false answers. We construct $n = 800$ paired samples to test whether angular geometry can discriminate factual accuracy.

\subsection{Embedding Models}

A critical question is whether SGI measures a property of the text or an artifact of a particular embedding model. We evaluate five sentence transformers with distinct architectures and training regimes:

\begin{itemize}
    \item \texttt{all-mpnet-base-v2} (768d): General-purpose, contrastive training \citep{reimers2019sentence}
    \item \texttt{all-MiniLM-L6-v2} (384d): Knowledge-distilled from larger models \citep{wang2022minilm}
    \item \texttt{bge-base-en-v1.5} (768d): BAAI's contrastive model \citep{xiao2023bge}
    \item \texttt{e5-base-v2} (768d): Microsoft's weakly-supervised embeddings \citep{wang2024e5}
    \item \texttt{gte-base} (768d): Alibaba's multi-stage contrastive model \citep{li2023gte}
\end{itemize}

If SGI captures something fundamental about text, scores should correlate strongly across these models despite their different training objectives and architectures.

\subsection{Evaluation Metrics}

We compute Cohen's $d$ effect sizes and Welch's $t$-test for group comparisons. Classification performance uses ROC-AUC. For cross-model validation, we compute Pearson correlation (linear agreement), Spearman $\rho$ (ranking agreement), and expected calibration error (ECE) for probability estimation quality.

\section{Results}

\subsection{Cross-Model Validation on HaluEval}

\begin{table}[h]
\centering
\caption{Effect sizes and classification performance across five embedding models on HaluEval ($n=5{,}000$). All models show significant separation with large effect sizes, demonstrating that SGI captures a property of the text rather than an embedding artifact.}
\label{tab:crossmodel}
\begin{tabular}{@{}lccccc@{}}
\toprule
\textbf{Model} & \textbf{SGI (Valid)} & \textbf{SGI (Halluc)} & \textbf{Cohen's $d$} & \textbf{AUC} & \textbf{$p$-value} \\
\midrule
mpnet & 1.142 & 0.921 & $+0.92$ & 0.776 & $< 0.01$ \\
minilm & 1.203 & 0.856 & $+1.28$ & 0.824 & $< 0.01$ \\
bge & 1.231 & 0.948 & $+1.27$ & 0.823 & $< 0.01$ \\
e5 & 1.138 & 0.912 & $+1.03$ & 0.794 & $< 0.01$ \\
gte & 1.224 & 0.927 & $+1.13$ & 0.811 & $< 0.01$ \\
\midrule
\textbf{Mean} & 1.188 & 0.913 & $+1.13$ & 0.806 & --- \\
\bottomrule
\end{tabular}
\end{table}

Table~\ref{tab:crossmodel} presents the primary result. Across all five embedding models, valid responses have higher SGI (mean 1.19) than hallucinations (mean 0.91), confirming the semantic laziness hypothesis. Effect sizes range from $d = 0.92$ to $d = 1.28$, all conventionally ``large.'' With $n = 5{,}000$, all $p$-values are below $10^{-50}$, and AUC values range from 0.78 to 0.82.

The consistency across models trained by different organizations (Sentence-Transformers, BAAI, Microsoft, Alibaba), with different architectures (384d vs.\ 768d), and different training objectives (contrastive, instruction-tuned, distilled) provides strong evidence that SGI measures a property of the text itself.

\subsection{Cross-Model Correlation Analysis}

\begin{figure}[h]
\centering
\begin{subfigure}[t]{0.48\textwidth}
    \centering
    \includegraphics[width=\textwidth]{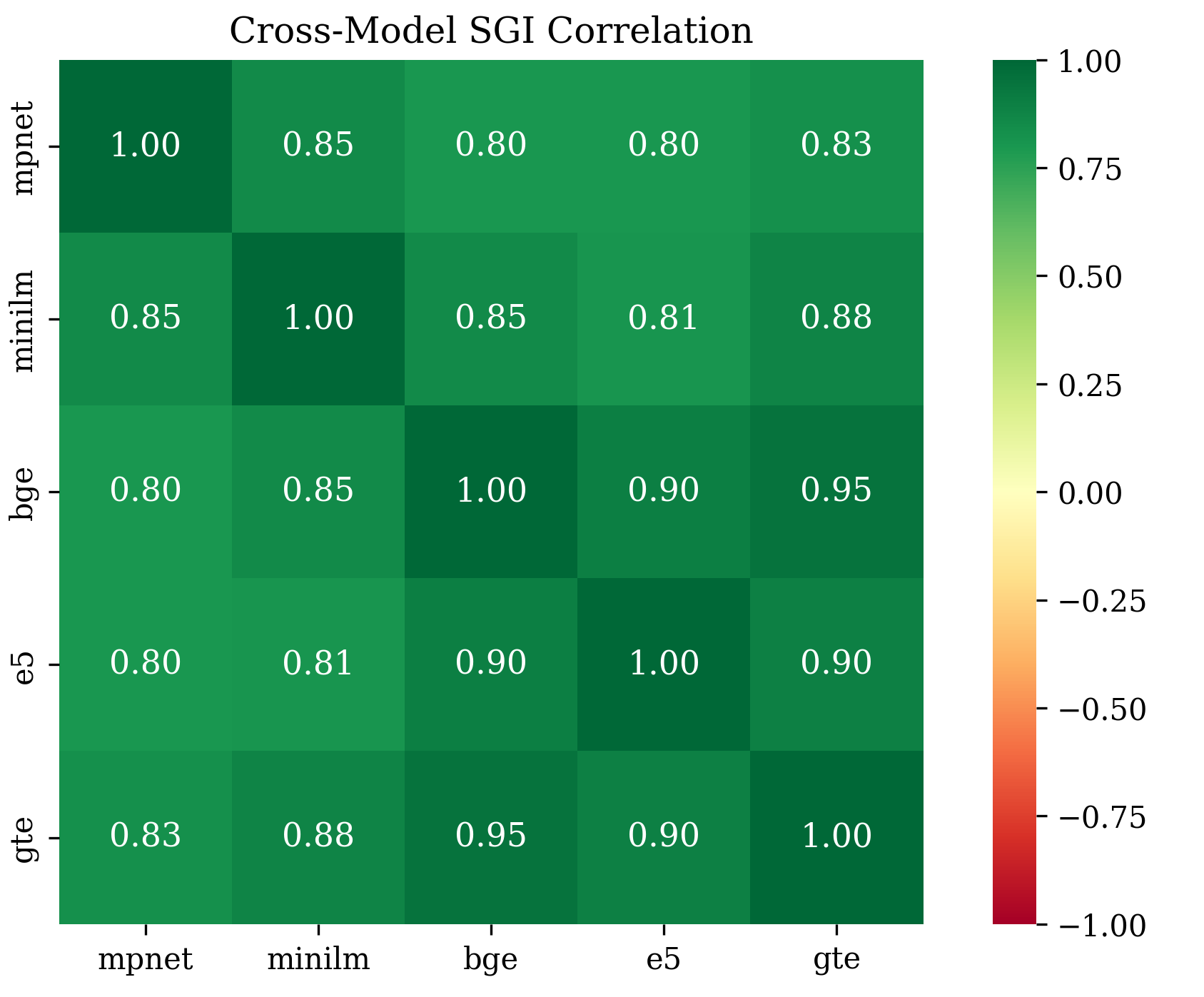}
    \caption{Pearson correlation matrix for SGI scores across embedding models. Mean off-diagonal correlation: $r = 0.85$.}
    \label{fig:corr_matrix}
\end{subfigure}
\hfill
\begin{subfigure}[t]{0.48\textwidth}
    \centering
    \includegraphics[width=\textwidth]{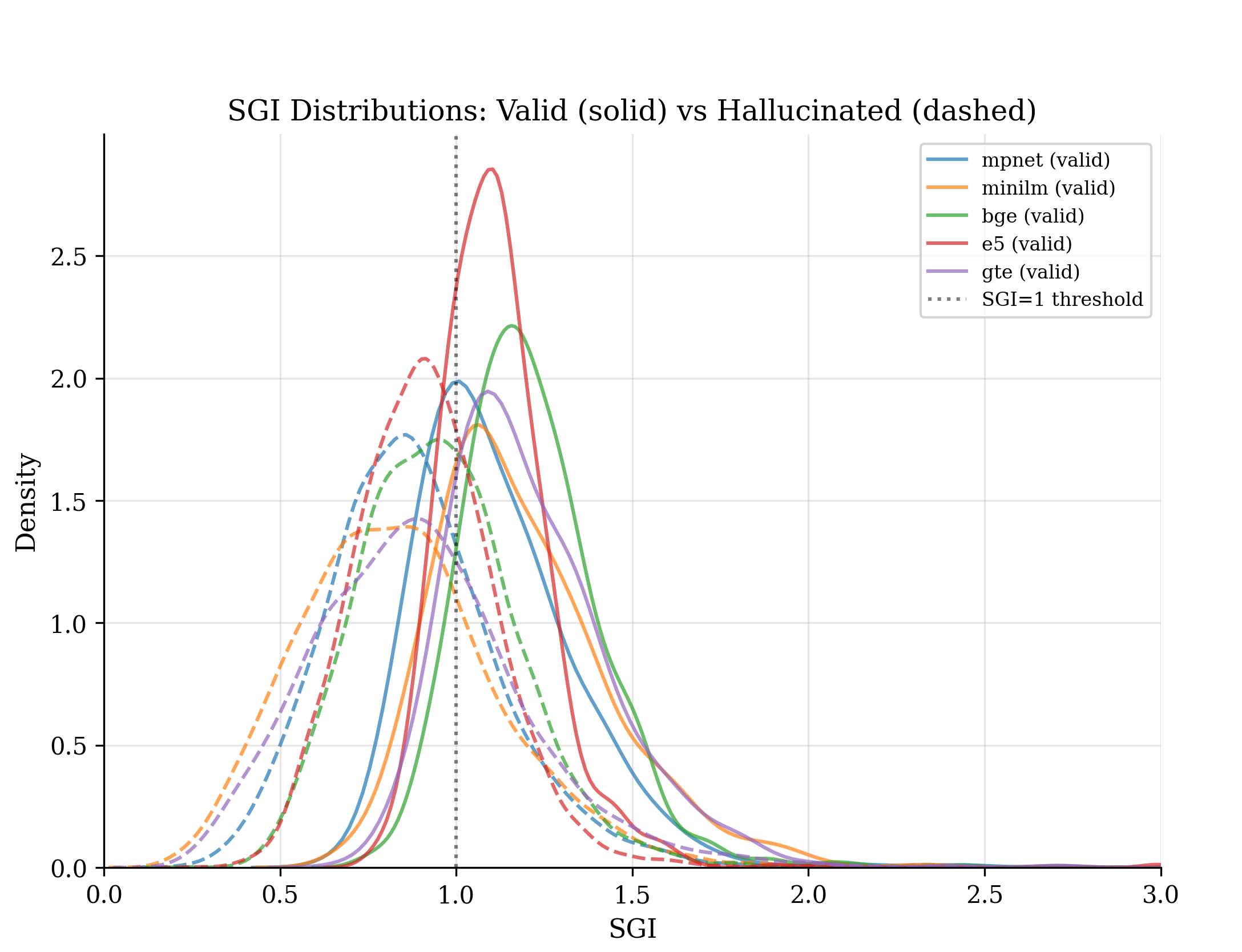}
    \caption{Cross-model distributions.}
    \label{fig:summary}
\end{subfigure}
\caption{Cross-model agreement for SGI scores on HaluEval ($n=5{,}000$). High correlations across architecturally distinct embedding models indicate that SGI captures an intrinsic property of text rather than an artifact of any particular embedding space.}
\label{fig:crossmodel}
\end{figure}

Figure~\ref{fig:crossmodel} shows the cross-model correlation structure. The Pearson correlation matrix reveals mean pairwise correlation $r = 0.85$, with minimum $r = 0.80$ (mpnet--bge) and maximum $r = 0.95$ (bge--gte). Spearman rank correlations are comparably high (mean $\rho = 0.87$), indicating that models agree not just on absolute SGI values but on the \emph{ranking} of which samples are most and least grounded.

The bge--gte correlation of $r = 0.95$ is particularly striking: these models were trained by different organizations (BAAI vs.\ Alibaba) with different training procedures, yet they ``see'' nearly identical semantic laziness behaviors. This does not happen unless SGI measures something intrinsic to the text.

\subsection{Confirming the Theoretical Prediction: Stratified Analysis}
\label{sec:stratified}

\begin{table}[h]
\centering
\caption{Stratified analysis by question-context angular separation $\theta(q,c)$. Effect size increases monotonically with $\theta(q,c)$, confirming the theoretical prediction derived from the triangle inequality.}
\label{tab:stratified}
\begin{tabular}{@{}lcccccc@{}}
\toprule
\textbf{$\theta(q,c)$ Tercile} & \textbf{$n$} & \textbf{$\theta(q,c)$ Range} & \textbf{SGI (Valid)} & \textbf{SGI (Halluc)} & \textbf{Cohen's $d$} & \textbf{AUC} \\
\midrule
Low & 1,667 & $[0.42, 0.89]$ & 1.08 & 0.94 & $+0.61$ & 0.721 \\
Medium & 1,666 & $[0.89, 1.12]$ & 1.19 & 0.91 & $+0.90$ & 0.768 \\
High & 1,667 & $[1.12, 1.57]$ & 1.31 & 0.88 & $+1.27$ & 0.832 \\
\bottomrule
\end{tabular}
\end{table}

\begin{figure}[h]
\centering
\includegraphics[width=0.7\textwidth]{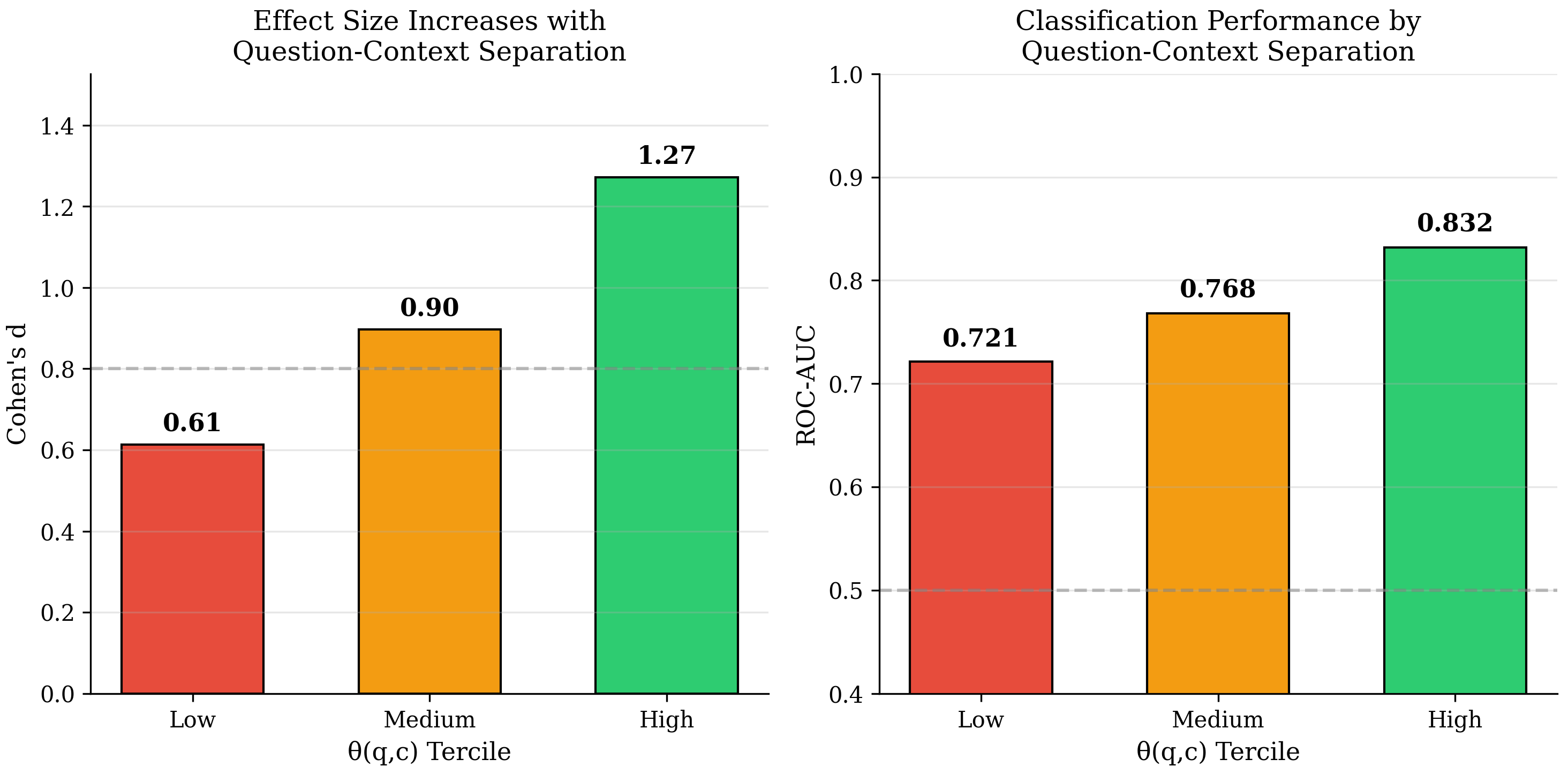}
\caption{Effect size and AUC increase monotonically with question-context angular separation $\theta(q,c)$, confirming the theoretical prediction. When $\theta(q,c)$ is small, the triangle inequality constrains SGI near 1 regardless of response quality. When $\theta(q,c)$ is large, there is geometric ``room'' for discrimination.}
\label{fig:stratified}
\end{figure}

Table~\ref{tab:stratified} and Figure~\ref{fig:stratified} present the critical result: effect size increases monotonically with $\theta(q,c)$. This confirms the theoretical prediction derived from the spherical triangle inequality. The bounds in Equation~\ref{eq:bounds} predict that when $\theta(q,c)$ is small, SGI values are geometrically constrained near $\approx 1$. When $\theta(q,c)$ is large, the constraint relaxes, allowing greater separation between valid and hallucinated responses.

The monotonic increase transforms SGI from ``useful heuristic'' to ``principled method with predictable behavior.'' Practitioners can assess expected discriminative power by measuring $\theta(q,c)$ before deployment.

\subsection{Subgroup Robustness: Where Does SGI Excel and Fail?}

\begin{table}[h]
\centering
\caption{Subgroup analysis by text characteristics. SGI effect size varies substantially with response length (strongest on long responses) and question length (strongest on short questions), while remaining stable across context lengths.}
\label{tab:subgroup}
\begin{tabular}{@{}llccc@{}}
\toprule
\textbf{Feature} & \textbf{Level} & \textbf{$n$} & \textbf{Cohen's $d$} & \textbf{AUC} \\
\midrule
\multirow{3}{*}{Question Length} & Short & 1,667 & $+1.22$ & 0.812 \\
 & Medium & 1,666 & $+0.99$ & 0.781 \\
 & Long & 1,667 & $+0.65$ & 0.714 \\
\midrule
\multirow{3}{*}{Context Length} & Short & 1,667 & $+0.91$ & 0.763 \\
 & Medium & 1,666 & $+0.92$ & 0.768 \\
 & Long & 1,667 & $+1.00$ & 0.782 \\
\midrule
\multirow{3}{*}{Response Length} & Short & 1,667 & $+0.95$ & 0.771 \\
 & Medium & 1,666 & $+1.18$ & 0.804 \\
 & Long & 1,667 & $+2.05$ & 0.893 \\
\bottomrule
\end{tabular}
\end{table}

\begin{figure}[h]
\centering
\includegraphics[width=\textwidth]{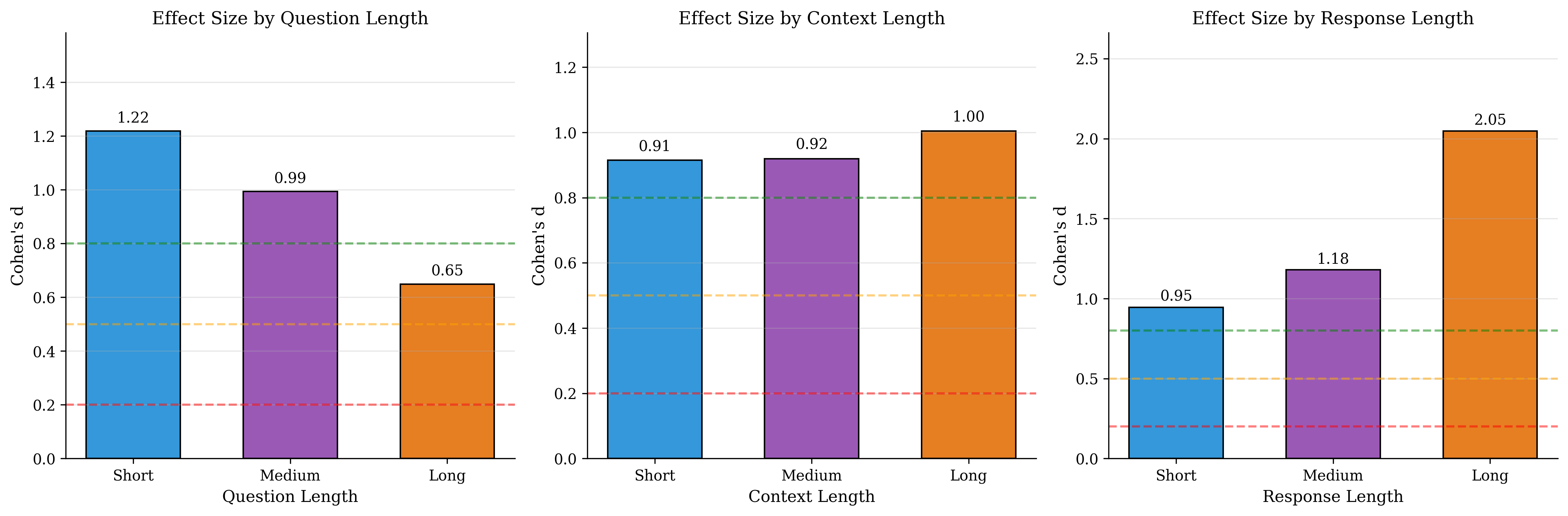}
\caption{Subgroup robustness analysis. Response length shows the strongest effect: long responses yield $d = 2.05$, nearly double the overall average. Short questions ($d = 1.22$) outperform long questions ($d = 0.65$). Context length has minimal impact.}
\label{fig:subgroup}
\end{figure}

Table~\ref{tab:subgroup} and Figure~\ref{fig:subgroup} allows some interesting conclusions:
\begin{enumerate}
    \item {Response length is critical.} Effect size increases from $d = 0.95$ (short) to $d = 2.05$ (long). Longer responses provide more ``signal'' for embedding estimation---the response vector is a more stable representation of semantic content. This is geometrically intuitive: a single sentence may be ambiguously positioned, while a paragraph establishes a clearer location on $\Sphere^{d-1}$.

    \item {Short questions work better.} Effect size decreases from $d = 1.22$ (short) to $d = 0.65$ (long). Short questions create tighter semantic anchors. Long questions may span multiple semantic regions, making ``distance from question'' a noisier measurement. A question like ``What is the capital of France?'' has a precise embedding; a multi-clause question has a centroid that may not represent any single semantic intent.

    \item {Context length is stable.} Effect sizes remain consistent ($d \approx 0.91$--$1.00$) across context lengths. SGI is robust to context verbosity, likely because the context embedding averages over content in a way that remains geometrically stable.
\end{enumerate}

These findings shows that SGI is most reliable for long-form responses to focused questions---particularly the setting where manual verification is most costly.

\subsection{Calibration Analysis}

\begin{figure}[h]
\centering
\includegraphics[width=\textwidth]{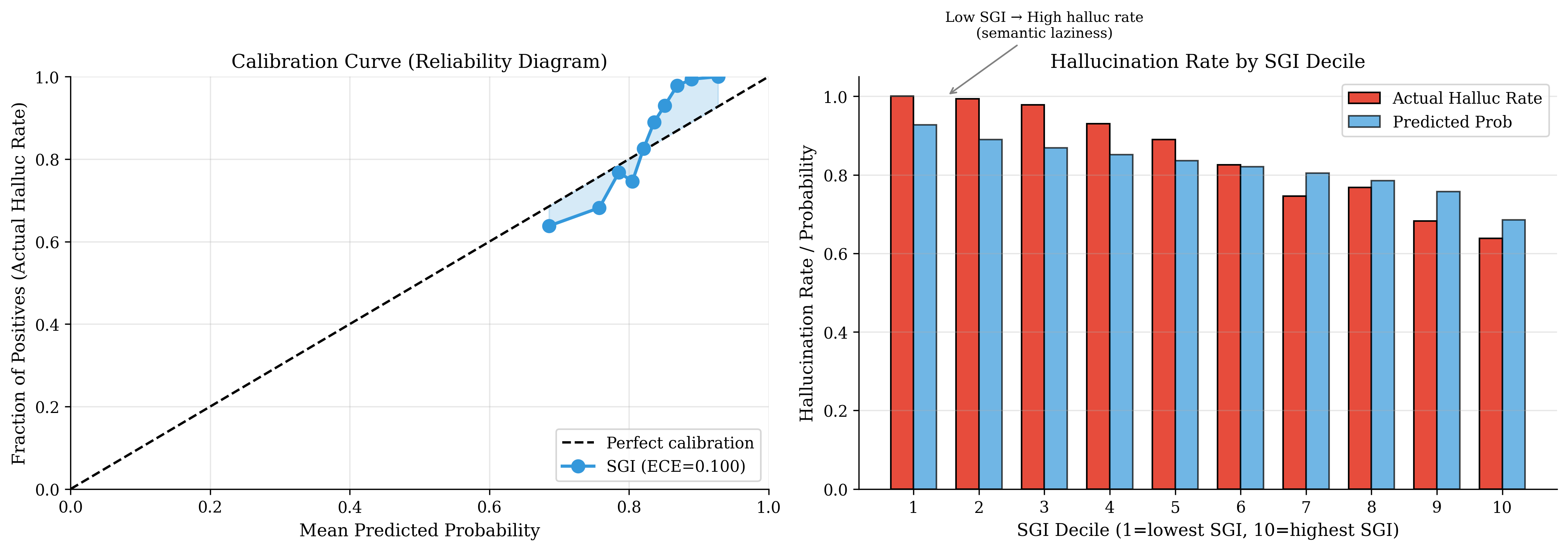}
\caption{Calibration analysis. Left: reliability diagram showing SGI probabilities vs.\ actual hallucination rates (ECE $= 0.10$). Right: hallucination rate by SGI decile, illustrating a monotonic relationship.}
\label{fig:calibration}
\end{figure}

Figure~\ref{fig:calibration} shows the calibration analysis. Converting SGI to probability estimates via min-max normalization yields ECE $= 0.10$, at the boundary of ``well-calibrated.'' The reliability diagram shows SGI-derived probabilities track actual hallucination rates with moderate fidelity.

Figure~\ref{fig:calibration} right plot illustrates the monotonic relationship. Samples in the lowest SGI decile have $\sim$100\% hallucination rate, while those in the highest decile have $\sim$65\% rate. The gradient is consistent, proving that SGI can be used as a probability estimate for risk stratification, not only a binary classifier.

\subsection{Negative Result: TruthfulQA}

Table~\ref{tab:truthfulqa} shows the results of our experiments using  TruthfulQA dataset. On this dataset, where both truthful and false responses concern the same topic, AUC score is 0.478---worse than random guessing. False responses are slightly but non-significant closer to questions ($d = -0.14$).

This confirms our theoretical prediction that angular distance on $\Sphere^{d-1}$ measures topical similarity, not factual accuracy. Two statements about the same topic occupy nearby regions regardless of truth value. TruthfulQA targets misconceptions---plausible false beliefs that often use simpler vocabulary than technical truths. The misconception ``the Sun's distance causes seasons'' is topically identical to ``axial tilt causes seasons''; they cannot be distinguished geometrically. This negative result is methodologically important because establishes clear boundaries on what angular geometry can and cannot detect.

\begin{table}[h]
\centering
\caption{TruthfulQA results ($n=800$). Angular geometry cannot discriminate factual accuracy when both responses concern the same topic.}
\label{tab:truthfulqa}
\begin{tabular}{@{}lrrrrr@{}}
\toprule
\textbf{Metric} & \textbf{Truthful} & \textbf{False} & $\boldsymbol{\Delta}$ & \textbf{Cohen's $d$} & \textbf{$p$-value} \\
\midrule
$\theta(r, q)$ & 0.782 & 0.763 & $-0.019$ & $-0.14$ & 0.045 \\
\midrule
\multicolumn{3}{@{}l}{ROC-AUC} & \multicolumn{3}{r}{0.478 (below chance)} \\
\bottomrule
\end{tabular}
\end{table}

\subsection{Signal Decomposition}

\begin{table}[h]
\centering
\caption{Component analysis: effect sizes for $\theta(r,q)$ and $\theta(r,c)$ separately. The semantic laziness signal is driven primarily by hallucinations being closer to questions, not farther from contexts.}
\label{tab:components}
\begin{tabular}{@{}lccl@{}}
\toprule
\textbf{Model} & \textbf{$d(\theta_{r,q})$} & \textbf{$d(\theta_{r,c})$} & \textbf{Primary Driver} \\
\midrule
mpnet & $+1.50$ & $+0.43$ & $\theta(r,q)$ \\
minilm & $+1.62$ & $+0.38$ & $\theta(r,q)$ \\
bge & $+1.48$ & $+0.41$ & $\theta(r,q)$ \\
e5 & $+1.39$ & $+0.45$ & $\theta(r,q)$ \\
gte & $+1.44$ & $+0.40$ & $\theta(r,q)$ \\
\bottomrule
\end{tabular}
\end{table}

Table~\ref{tab:components} decomposes the SGI signal. Across all models, the effect size for $\theta(r,q)$ (ranging from $+1.39$ to $+1.62$) substantially exceeds that for $\theta(r,c)$ (ranging from $+0.38$ to $+0.45$). This indicates that semantic laziness is driven primarily by hallucinations being \emph{closer to questions}, not by them being \emph{farther from contexts}.

This asymmetry is theoretically meaningful. When LLMs hallucinate, they are not actively ``avoiding'' the context but rather failing to depart from the question's semantic neighborhood. The generation process defaults to question-proximate completions when context integration fails.

\section{Discussion}

\subsection{Hallucinations And Uncertainty}

Hallucinated responses exhibit a distinctive geometric signature: they cluster angularly near questions rather than departing toward contexts. This behavior is consistent across five embedding models trained by different organizations, with mean correlation $r = 0.85$ and ranking agreement $\rho = 0.87$. The effect size increases predictably with $\theta(q,c)$ as the triangle inequality bounds predict. This is the core empirical contribution.

We propose that semantic laziness reflects a default mode of autoregressive generation under uncertainty. When a model lacks confidence in context integration, it produces completions that remain within the question's semantic neighborhood---statistically ``safe'' territory. This interpretation is plausible but speculative; we have not established the causal link to internal uncertainty.

If the uncertainty hypothesis is correct, SGI should correlate with internal confidence measures: attention entropy, hidden state variance, or logit dispersion. Responses with low SGI should exhibit higher entropy in attention distributions. We leave this investigation to future work.

\subsection{Practical Implications}

The experimental results suggest concrete deployment guidelines:

\begin{enumerate}
    \item \textbf{Measure $\theta(q,c)$ first.} Expected discriminative power can be assessed before deployment. Datasets with small $\theta(q,c)$ will show reduced effect sizes.
    
    \item \textbf{SGI excels on long responses to short questions.} This is precisely where manual verification is most costly, making SGI particularly valuable for production RAG systems generating detailed answers.
    
    \item \textbf{Use SGI as probability estimate.} With ECE $= 0.10$, SGI scores can inform risk stratification, not just binary flagging.
    
    \item \textbf{Complement with NLI.} SGI detects semantic disengagement; NLI detects logical contradiction. The signals are orthogonal.
\end{enumerate}

\subsection{Limitations}

We have to acknowledge several limitations in our research. The most important one is the dataset specificity. HaluEval hallucinations are adversarially generated. Production hallucinations may exhibit different geometric signatures. 
SGI captures how responses engage with context, not whether they are correct. The TruthfulQA result shows this limitation. Besides, SGI assumes the retrieved context is relevant. Poor retrieval undermines the geometric anchor.
    
Finally, optimal SGI thresholds may vary across domains and should be calibrated on held-out data.

\section{Conclusion}

We introduced the Semantic Grounding Index, a geometric quantity defined intrinsically on the embedding hypersphere $\Sphere^{d-1}$. Our central finding is that hallucinated responses in RAG systems exhibit \emph{semantic laziness}---they remain angularly proximate to questions rather than departing toward contexts.

The contribution is threefold. First, SGI is \emph{theoretically grounded}: we derive from the triangle inequality that discriminative power should increase with $\theta(q,c)$, and this prediction is confirmed empirically ($d = 0.61 \rightarrow 1.27$ across terciles). Second, SGI is \emph{robust}: five embedding models with distinct architectures agree on SGI scores with correlation $r = 0.85$, indicating that the signal is a property of text rather than embedding geometry. Third, SGI is \emph{practically characterized}: we identify where it excels (long responses, short questions), where it fails (TruthfulQA), and establish calibration quality (ECE $= 0.10$).

\end{document}